\documentclass{article}
\usepackage{graphicx}
\usepackage{float}
\usepackage{wrapfig}
\usepackage[preprint]{neurips_2026}

\usepackage[utf8]{inputenc} 
\usepackage[T1]{fontenc}    
\usepackage{hyperref}       
\usepackage{url}            
\usepackage{booktabs}       
\usepackage{amsfonts}       
\usepackage{nicefrac}       
\usepackage{microtype}      
\usepackage{xcolor}         
\usepackage{subcaption}

\usepackage{xspace}
\usepackage{amsmath}

\newcommand{\squishlist}{
   \begin{list}{\small $\bullet$}
    { \setlength{\itemsep}{0pt}      \setlength{\parsep}{4pt}
      \setlength{\topsep}{1pt}       \setlength{\partopsep}{4pt}
     \setlength{\leftmargin}{1.2em} \setlength{\labelwidth}{1em}
      \setlength{\labelsep}{0.5em} } }
\newcommand{\squishend}{  \end{list}  }

\newcommand{\xhdr}[1]{\noindent\textbf{#1:}\xspace}

\title{Early Prediction of Future Behavioral Strategy from Process Traces}

\author{%
  Robert Kasumba  \\
  Division of Computational and Data Sciences\\
  Washington University in Saint Louis\\
  Saint Louis, MO 63130 \\
  \texttt{rkasumba@wustl.edu} \\
    \And 
Dennis Barbour  \\
  Department of Biomedical Engineering\\
  Washington University in Saint Louis\\
  Saint Louis, MO 63130 \\
  \texttt{dbarbour@wustl.edu} \\
  \And
  Chien-Ju Ho  \\
  Department of Computer Science\\
  Washington University in Saint Louis\\
  Saint Louis, MO 63130 \\
  \texttt{chienju.ho@wustl.edu} \\
}

\begin{document}

\maketitle

\begin{abstract}
Adaptive systems often need to make task-specific decisions about people from limited evidence: a tutor may need to anticipate how a learner will approach a new problem, a game may need to adapt when a player enters a new level, and a human-AI system may need to infer whether a partner will persist with a plan or switch goals. These decisions depend on person-level tendencies that shape how people solve related tasks, but such tendencies are difficult to infer from standard behavioral evidence. One approach is to use aggregate outcome summaries, such as scores, completion rates, or productivity; these summaries are compact and available across tasks, but can collapse distinct behavioral processes into similar outcomes. Another approach is to use process-level traces, which record how behavior unfolds; however, process modeling within one task can entangle stable person-level tendencies with task-specific layout and affordances. In this work, we study early cross-task behavioral inference: whether partial source-task process traces can reveal transferable person-level structure that predicts strategy in a held-out target task. We introduce a Process-Level Latent Variable Model (PLVM), which encodes task-specific traces and fuses them into a shared person-level latent representation for cross-task prediction. In PowerWash Simulator, a naturalistic telemetry dataset of human gameplay, PLVM uses partial traces from two cleaning tasks to predict locally persistent Zone Planner behavior versus frequent Zone Hopper behavior in the held-out Fire Station level. Controlled simulations with known latent types show that cross-task fusion helps when source tasks reveal complementary dimensions of a shared latent process. These results suggest that process-level cross-task modeling can support early prediction of target-task strategy when observing sufficient target-task behavior is impractical.
\end{abstract}
\section{Introduction}

Adaptive systems often need to make task-specific decisions about a person from limited prior evidence. A tutor may need to anticipate how a learner will approach a new problem \cite{vanlehn2011relative}, a game may need to adapt when a player enters a new level\cite{paraschos2023game}, and a human-AI partner may need to infer whether a collaborator will persist with a plan or switch goals in a new collaborative episode~\cite{hoffman2024inferring, kasumbagoal}. Such decisions depend on person-level behavioral tendencies that shape how people solve related tasks. The useful prediction is therefore not only what outcome the person has achieved before, but how the person is likely to organize behavior in the target task.

A natural starting point for this prediction problem is to use outcome summaries of prior behavior. Behavioral science and applied assessment have often relied on measures such as accuracy, reaction time, score, completion time, or total progress to estimate individual differences \cite{miyake2000unity,friedman_unity_2017,kasumba2023distributional, kasumba2025bayesian}. These summaries are compact and often available across tasks, but they can underdetermine the behavioral process that produced them: similar outcomes can arise from systematic spatial organization, local exploitation, frequent target switching, or long-range movement toward future subtasks. Even when outcome summaries are combined across tasks, they abstract away how behavior unfolds within each task \cite{miyake2000unity,friedman_unity_2017,marticorena2024minecraft}. This abstraction is limiting for adaptive systems because different behavioral processes may call for different interventions even when observed outcomes are similar.

This motivates using process traces: time-ordered behavioral records that preserve information outcome summaries discard \cite{jain2023computational, anghel2024use}. In our setting, these traces capture how behavior unfolds within a task, including where people move, when they switch targets, how they approach subtasks, and how progress accumulates. Modern sequence models, such as transformers \cite{vaswani2017attention}, can learn rich temporal dependencies in such traces and model how behavior continues within a task. However, richer process data alone does not solve the cross-task inference problem. A strong sequence model may learn the layout, timing, and local affordances of a particular environment, entangling task-specific regularities with stable person-level tendencies. Conversely, a relevant person-level tendency may be only partially expressed in any single source task: a learner's persistence, a player's search style, or a teammate's switching tendency may appear differently across environments. The key challenge is therefore not merely to model process traces within a task, but to combine process evidence across tasks to infer behavioral structure that transfers to a held-out target task.

In this work, we study early cross-task behavioral inference: predicting how a person will organize behavior in a held-out target task from partial process traces in prior source tasks. We ask whether process traces from multiple source tasks can be fused to recover transferable person-level behavioral structure beyond what is captured by aggregate outcome summaries or single-task process models. We instantiate this problem in the PowerWash Simulator gameplay dataset~\cite{vuorre2023intensive}, which contains naturalistic telemetry from human players completing long-horizon spatial cleaning tasks. We use partial traces from two source levels to predict navigation style in a later held-out target level: whether a player tends to persist locally within spatial regions or repeatedly switch across regions. This distinction is actionable because players with similar completion outcomes may require different forms of assistance, feedback, or coordination depending on how they organize their behavior.

We propose a Process-Level Latent Variable Model (PLVM). PLVM uses task-specific encoders to summarize process traces from each source task, then fuses the summaries into a shared person-level latent representation used to predict behavior in the held-out target task. The goal is not to continue a sequence within one task, but to infer transferable behavioral organization that may be expressed differently across tasks. Because real gameplay does not reveal the true latent process underlying behavior, we use controlled simulations as a mechanism check: simulated agents have known latent behavioral types, and different source tasks reveal complementary aspects of those types. In PowerWash Simulator, PLVM uses partial traces from two source cleaning tasks to predict whether a player will later behave as a locally persistent Zone Planner or a frequent Zone Hopper in a held-out target level. PLVM outperforms outcome-based models, which remain close to random baseline performance, and consistently improves over single-task process models. These results suggest that transferable behavioral strategy is not fully captured by aggregate outcomes or by process traces from either source task alone. In controlled simulations, PLVM better recovers ground-truth behavioral phenotypes when different source tasks reveal complementary aspects of the same underlying process. Overall, our work makes three contributions:
\squishlist
    \item We formulate early cross-task behavioral strategy prediction as an application-relevant problem: using partial process traces from source tasks to predict behavioral strategy in a held-out task.
    \item We introduce PLVM, a process-level latent representation model that fuses task-specific trace embeddings into a shared person-level representation for cross-task prediction.
    \item We show in both real human gameplay and controlled simulation that cross-task process modeling can improve prediction of target-task behavioral strategy when aggregate outcome summaries and single-task process models are insufficient.
\squishend
\section{Related Work}

\xhdr{Process tracing and behavioral assessment}
Cognitive science has long sought to infer latent cognitive processes from observable behavior. Traditional assessment often relies on outcomes such as accuracy, reaction time, completion time, or total score to estimate individual differences in constructs such as working memory, attention, or cognitive flexibility \cite{miyake2000unity,friedman_unity_2017}. These measures are useful but can collapse distinct behavioral processes: two people may achieve similar performance while differing in how they search, plan, persist, or switch goals \cite{marticorena2024minecraft}. Process-tracing methods address this limitation by treating fine-grained behavioral traces as evidence about latent planning and decision strategies \cite{callaway2017mouselab,jain2023computational}. This work shows that process data can reveal structure that outcome summaries miss. However, much of this literature focuses on explaining behavior within a task or estimating strategy from traces in the same setting. Our problem is different: we ask whether partial process traces from prior tasks can predict how an individual will organize behavior in a future task.

\xhdr{Search, foraging, and navigation style}
Our target behavioral label is motivated by theories of search and foraging. Foraging models study how agents search for distributed resources, exploit local patches, decide when to leave a region, and trade off reward against movement or travel costs \cite{charnov1976optimal}. Related ideas appear in information foraging and visual foraging, where human search behavior is shaped by local persistence, switching, and the expected value of moving to another region \cite{pirolli1999information,bella2022foraging,kristjansson2020dynamics}. These theories provide a useful vocabulary for describing spatial behavior in long-horizon tasks: some individuals persist locally, whereas others switch more frequently across regions. In our work, this distinction motivates the Zone Planner versus Zone Hopper target in PowerWash. The contribution is not a new theory of foraging, but a predictive modeling framework that learns whether such behavioral organization transfers across tasks from partial process traces.

\xhdr{Behavioral telemetry, virtual environments, and individual style}
Games and virtual interactive environments provide dense, naturalistic behavioral traces at scale, making them useful for studying individual differences in complex behavior \cite{marticorena2024minecraft}. Prior work on behavioral telemetry has used such traces to model performance, engagement, skill, well-being, or decision style. Closely related is behavioral stylometry, which uses patterns of decisions to detect individual decision-making style, for example in chess \cite{mcilroy2021detecting}. This literature demonstrates that individual style can be measured from behavioral sequences. However, stylometry and player modeling often operate within a single domain or aim to identify individuals, predict outcomes, or characterize behavior after observation. We instead study an early cross-task prediction problem: given partial behavior in source tasks, predict behavioral strategy in a held-out target task whose traces are not observed as input.

\xhdr{Sequence models and cross-task representation learning}
Sequence models, including transformers, can learn temporal dependencies from behavioral traces \cite{vaswani2017attention}. They are well suited for modeling within-task dynamics, such as how recent actions predict future actions or how local progress predicts subsequent movement. However, strong within-task prediction does not guarantee recovery of transferable person-level structure. A sequence model may exploit the affordances of a particular task while entangling those regularities with individual behavioral tendencies. Multi-task and domain representation learning address related issues by sharing information across tasks or learning features that generalize across domains, but they are not usually framed around person-level behavioral inference from multiple partial traces \cite{yu2025multitask,collins2024provable,khoee2024domain,lippl2024inductive}. PLVM addresses this setting directly: task-specific encoders summarize process traces from each source task, and a fusion module learns a shared person-level representation used to predict behavior in a held-out target task.

\section{Methods}

\subsection{Problem Formulation}
We consider individuals $i \in \{1,\ldots,N\}$ observed across multiple tasks. Let $\mathcal{K}_{obs}$ denote the set of observed source tasks and let $k^\star$ denote a held-out target task. For each individual $i$ and source task $k \in \mathcal{K}_{obs}$, we observe a partial process trace
\[
X_{i,k}^{(t)} = (x_{i,k,1}, \ldots, x_{i,k,t}),
\]
where each token records process-level information available at event $t$, such as position, progress, elapsed time, local task state, or subtask-completion information. The goal is to predict a behavioral target $y_{i,k^\star}$ in the held-out target task using only partial traces from the source tasks. 

\subsection{Process-Level Latent Variable Model (PLVM)}
\begin{figure}
    \centering
    \includegraphics[width=0.8\linewidth]{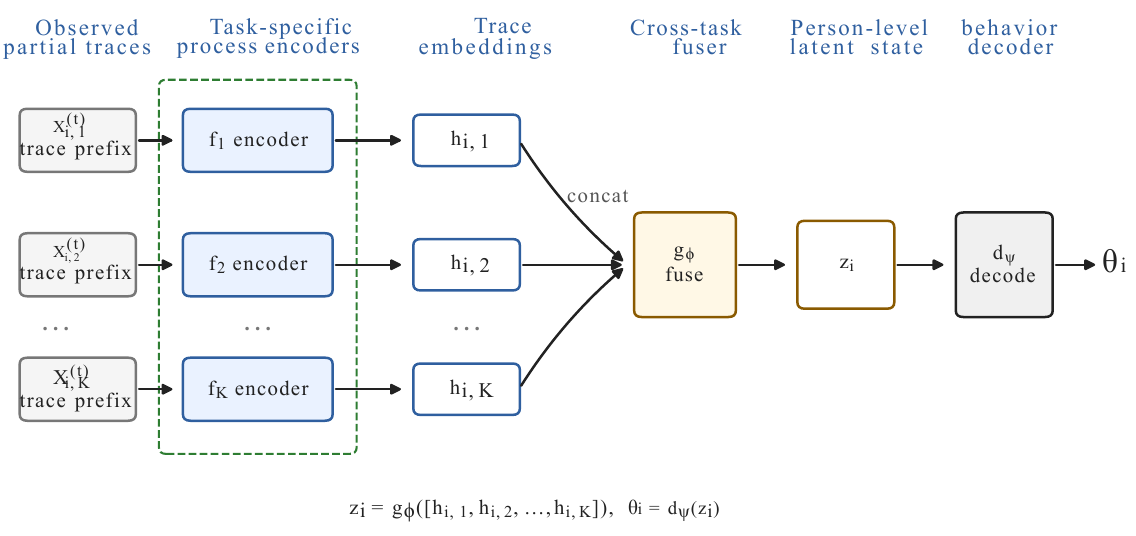}
    \caption{PLVM architecture. Task-specific process encoders map partial source-task traces to fixed-dimensional embeddings. A cross-task fuser combines these embeddings into a shared person-level latent representation, which is decoded to predict the behavioral target in a held-out target task. }
    \label{fig:architecture}
\end{figure}

PLVM consists of three components: task-specific process encoders, a cross-task fuser, and a target-behavior decoder. Each task encoder summarizes a partial process trace from one source task. The cross-task fuser combines these task embeddings into a shared person-level latent representation, and the decoder uses this representation to predict the behavioral target. In our implementation, the task-specific encoders are trained independently, then frozen; the fuser and decoder are trained jointly for the downstream cross-task prediction task. Figure~\ref{fig:architecture} shows the architecture.

\xhdr{Task-specific process encoders}
For each source task $k \in \mathcal{K}_{obs}$, we train an encoder $f_k$ on process traces from that task. The encoder maps a partial trace to a fixed-dimensional embedding:
\[
h_{i,k}^{(t)} = f_k(X_{i,k}^{(t)}).
\]
In our experiments, $f_k$ is a causal transformer trained with a self-supervised next-event prediction objective. This objective is not the final scientific target; it is used to learn representations of within-task behavioral dynamics. In simulation, the next-event target is the agent's next action or target choice. In PowerWash, the next-event target is the player's position change by the next subtask-completion event. The task-specific encoders are trained independently and then frozen; the fusion module and decoder are trained on the downstream cross-task prediction objective.

\xhdr{Cross-task fusion}
Given embeddings from the source tasks, PLVM constructs a shared person-level latent representation
\[
z_i^{(t)} = g_{\phi}\left(\operatorname{concat}_{k \in \mathcal{K}_{obs}} h_{i,k}^{(t)}\right),
\]
where $g_{\phi}$ is a feed-forward fusion network. The representation $z_i^{(t)}$ is intended to summarize behavioral structure that is shared across tasks. Because each task-specific encoder is trained separately, the fusion module learns how evidence from different source tasks should be combined for the downstream behavioral target.

\xhdr{Target-behavior decoder}
A decoder $d_{\psi}$ maps the person-level representation to prediction scores:
\[
\theta_i = d_{\psi}(z_i^{(t)}).
\]
For classification targets, such as behavioral phenotype in simulation or Zone Planner/Zone Hopper in PowerWash, these scores define
\[
p_{\psi}(y \mid z_i^{(t)}) = \operatorname{softmax}(\theta_i)_y.
\]
We train the fuser and decoder using cross-entropy loss:
\[
\mathcal{L}(\phi,\psi) =
-\sum_i \log p_{\psi}(y_{i,k^\star} \mid z_i^{(t)}).
\]
 $y_{i,k^\star}$ is defined from held-out target-task behavior, or from a known individual-level phenotype in simulation, while $z_i^{(t)}$ is computed only from partial traces in the source tasks.

\subsection{Model Evaluation}
\xhdr{Comparison models}
We compare PLVM to two classes of baselines. Outcome baselines use aggregate summaries of source-task behavior up to the same observation cutoff, such as completion, progress, reward, productivity, or recent-window statistics. These models test whether target-task behavioral strategy can be predicted from what the person accomplished, without modeling how behavior unfolded. Single-task process models use the same task-specific process encoders as PLVM but receive only one source task at a time. These models test whether detailed traces from a single task are sufficient without cross-task fusion. PLVM differs from both by combining process-level evidence from multiple source tasks into a shared person-level representation.

\xhdr{Partial observation protocol}
We evaluate all models under matched observation cutoffs. At cutoff $t$, each model receives only the prefix of each source-task trace up to that cutoff. Outcome baselines compute summaries only from the same prefix. Single-task process models and PLVM use only prefix embeddings from the task-specific encoders. The held-out target-task trace is never provided as input. This protocol tests whether a target-task behavioral strategy can be inferred from partial source-task behavior before complete source-task traces are available.

\section{Experiments}

We evaluate PLVM in two complementary ways. The primary experiment uses real human gameplay from PowerWash Simulator to test the application problem: predicting a player's future navigation style in a held-out task from partial traces in earlier tasks. The second experiment is a controlled simulation designed to test the mechanism: whether cross-task fusion helps when different tasks reveal complementary aspects of a known latent behavioral process.



\subsection{Evaluation Design}
Our experiments test three claims. First, future behavioral strategy should not be well predicted by aggregate early-task outcomes alone. Second, detailed process traces should improve prediction, but a single task may provide only a partial view of the person-level strategy. Third, fusing process evidence across tasks should improve prediction when the observed tasks reveal complementary aspects of the same behavioral organization. To test these claims, we compare PLVM with two alternatives. Outcome-based models use aggregate summaries of prior behavior, such as completion, productivity, and progress, to test whether future strategy can be predicted from what a person accomplished. Single-task sequence models use detailed process traces from one task at a time, testing whether within-task dynamics are sufficient without cross-task fusion. PLVM differs by encoding process traces from multiple tasks and fusing them into a shared person-level representation intended to capture transferable behavioral structure.

\subsection{PowerWash Simulator: Prediction of Future Navigation Style using  Human Gameplay }

\begin{figure}
    \centering
    \includegraphics[width=0.7\linewidth]{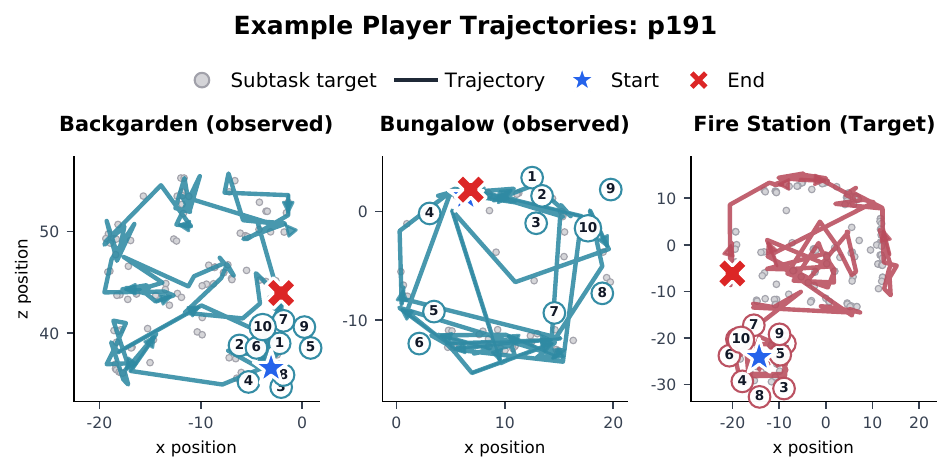}
    \caption{
Example PowerWash process traces across the three tasks we used.
Each panel shows subtask locations and one player's ordered completion trajectory.
Back Garden and Bungalow are used as observed tasks, while Fire Station is the future held-out target task.
}
    \label{fig:sample_powerwash_player}
    \vspace{-8pt}
\end{figure}

\xhdr{Dataset and Task Setup}
We use the \emph{PowerWash Simulator} gameplay dataset: real human telemetry from $11{,}080$ unique players completing long-horizon spatial cleaning tasks. These tasks require players to search for distributed targets, persist within local regions, switch areas, and move toward future subtask completions. We focus on three career-mode cleaning tasks: Back Garden, Bungalow, and Fire Station, which share the broad structure of spatial cleaning while differing in layout, scale, and local affordances. They also form a natural early-to-late career progression with substantial player overlap. Figure~\ref{fig:sample_powerwash_player} shows one player's behavioral trajectory across the three tasks. We use Back Garden and Bungalow as early observation tasks and Fire Station as the held-out future task: given only a player's process-level behavior in Back Garden and Bungalow, models predict the behavioral strategy the same player will use in Fire Station.

\begin{figure}
    \centering
    \begin{subfigure}[t]{0.28\linewidth}
        \centering
        \includegraphics[width=\linewidth]{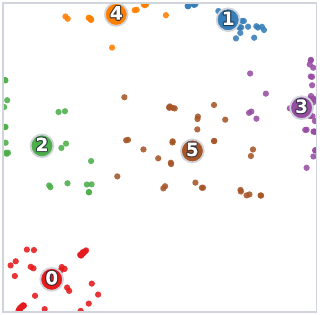}
        \caption{}
        \label{fig:sample_powerwash_player_back_garden}
    \end{subfigure}
    \hfill
    \begin{subfigure}[t]{0.28\linewidth}
        \centering
        \includegraphics[width=\linewidth]{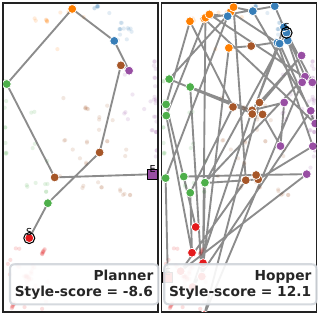}
        \caption{}
        \label{fig:sample_powerwash_player_bungalow}
    \end{subfigure}
    \hfill
    \begin{subfigure}[t]{0.28\linewidth}
        \centering
        \includegraphics[width=\linewidth]{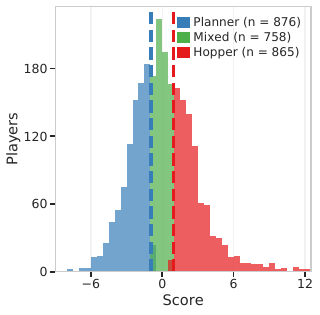}
        \caption{}
        \label{fig:label_distribution}
    \end{subfigure}

    \caption{
Construction of navigation-style labels. 
(a) Fire Station subtasks are grouped into six K-Means clusters(zones). 
(b) For each player, we convert the ordered subtask completions into a zone-completion trajectory; Zone Planners show more coherent local progression, whereas Zone Hoppers switch repeatedly across zones. 
(c) Players are labeled using the resulting navigation-style score: the bottom $35\%$ are labeled Planners, the top $35\%$ Hoppers, and the middle $30\%$ Mixed/ambiguous.
}
    \label{fig:zone_labeling}
\vspace{-8pt}
\end{figure}

\xhdr{Future Navigation-Style Target}
We operationalize future behavioral strategy as the tendency to switch between spatial regions. This is actionable since players with similar completion progress may require different forms of assistance: local object discovery support for persistent players versus route stabilization or reminders for frequent switchers. As shown in Figure~\ref{fig:zone_labeling}, we cluster Fire Station subtask locations into spatial zones and convert each player's subtask completions into a zone sequence $q_{i,1:T_i}$. We compute four summary statistics: zone switch rate $a_i$, return rate $r_i$, mean zone run length $\ell_i$, and mean zone jump distance $d_i$. The navigation-style score is
\[
S_i = z(a_i) + z(r_i) - z(\ell_i) + z(d_i),
\]
where $z(\cdot)$ denotes standardization across players. Higher scores indicate more frequent zone switching, more returns to previous zones, larger spatial transitions, and shorter same-zone runs; lower scores indicate locally persistent \emph{Zone Planners}, while higher scores indicate \emph{Zone Hoppers}. We label the bottom $35\%$ of $S_i$ as Zone Planners, the top $35\%$ as Zone Hoppers, and treat the middle $30\%$ as \emph{Mixed}/ambiguous; the primary binary task excludes this middle group.
\xhdr{Evaluation Protocol}
We compare PLVM with two classes of baselines. Outcome-based models test whether aggregate early-task behavior is sufficient to predict future strategy. Single-task transformers test whether detailed process traces are sufficient when each early task is modeled separately. PLVM tests whether fusing process-level embeddings across Back Garden and Bungalow yields a more transferable person-level representation. We evaluate all models by their ability to predict Fire Station behavioral strategy from early-task observations. An advantage over outcome-based models would indicate that dense process traces contain information not captured by aggregate summaries, while an advantage over single-task transformers would suggest that cross-task process modeling captures structure not fully recovered by within-task sequence dynamics alone.



\subsection{Controlled Grid-World Simulation : Mechanism Check}

\begin{figure}
    \centering
    \begin{subfigure}[t]{0.24\linewidth}
        \centering
        \includegraphics[width=\linewidth]{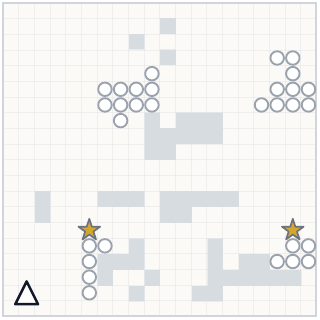}
        \caption{Foraging Task}
        \label{fig:Foraging_Layout}
    \end{subfigure}
    \hfill
    \begin{subfigure}[t]{0.24\linewidth}
        \centering
        \includegraphics[width=\linewidth]{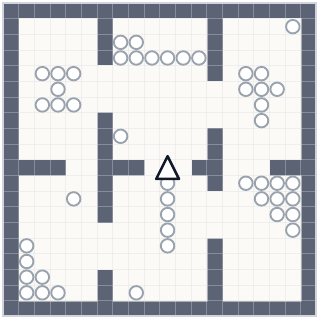}
        \caption{Cleaning Task}
        \label{fig:cleaning_layout}
    \end{subfigure}
    \hfill
    \begin{subfigure}[t]{0.24\linewidth}
        \centering
        \includegraphics[width=\linewidth]{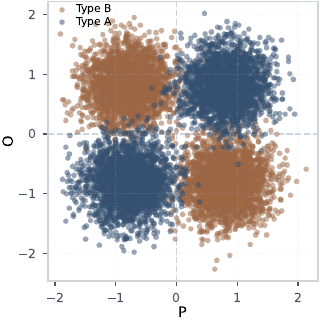}
        \caption{Simulated Agents}
        \label{fig:label_distribution}
    \end{subfigure}

    \caption{
Paired-task simulation setup and latent label structure. (a) Foraging environment: agents navigate a shared grid to collect food items (circles), including a small number of high-value targets (stars). (b) Cleaning environment: agents traverse a distinct grid layout to clear distributed mess targets organized into spatial patches. (c): The latent trait distribution used to generate player types.} 

    \label{fig:simulation_setup}
    \vspace{-8pt}
\end{figure}

\xhdr{Simulation Design}
We use a controlled grid-world simulation in which each synthetic agent completes two related tasks: an outdoor \emph{foraging} task and an indoor \emph{cleaning} task, shown in Figure~\ref{fig:simulation_setup}. In both tasks, agents complete spatially distributed targets, and movement to a selected target follows shortest-path navigation. Agents therefore differ not in pathfinding ability, but in which target they choose next and how those choices are organized over time.

\xhdr{Known Latent Phenotypes}
Each agent follows a simulated policy parameterized by two shared latent variables: patch persistence $P_i$, the tendency to remain within the same local region, and opportunistic reward seeking $O_i$, the tendency to prioritize salient or high-value targets. These variables influence target selection across both tasks, but with different task-specific strengths. Ground-truth labels are assigned by a diagonal rule: agents with same-sign drives, high $P_i$/high $O_i$ or low $P_i$/low $O_i$, are labeled \emph{Type A}; agents with opposite-sign drives, high $P_i$/low $O_i$ or low $P_i$/high $O_i$, are labeled \emph{Type B}. Thus, the phenotype depends on the joint configuration of the latent variables.
The tasks are designed to provide complementary evidence about this joint phenotype. In foraging, agents collect 32 food targets, including rare high-value items, making target choices more informative about $O_i$. In cleaning, agents clean 48 equal-valued targets arranged in clusters and strips, making choices more informative about $P_i$. At each decision point, agent $i$ selects the next unfinished target $j$ using a softmax over scores combining same-patch persistence, task-specific target value, shortest-path distance, and noise:
\[
\mathrm{score}_i^{(m)}(j)
=
\beta_P^{(m)} P_i \mathbf{I}\{j \in \text{ current\ patch}\}
+
\beta_O^{(m)} O_i V_j^{(m)}
-
\beta_D d(x_i, x_j)
+
\epsilon_j ,
\]
where $m$ indexes the task, $V_j^{(m)}$ is target value or salience, $d(x_i,x_j)$ is shortest-path distance from the agent to the target, and $\epsilon_j$ is stochastic choice noise. The coefficients $\beta_P^{(m)}$ and $\beta_O^{(m)}$ control how strongly persistence and opportunism are expressed in each task. After selecting a target, the agent takes the shortest path and completes it. This design makes every single task only partially informative: foraging provides stronger evidence about $O_i$, while cleaning provides stronger evidence about $P_i$. Since the phenotype depends on whether these two latent drives are aligned or conflicting, recovering it should benefit from combining evidence across both tasks. PLVM receives partial traces from both tasks and predicts the ground-truth phenotype. In contrast, single-task models receive traces from only one task, and outcome baselines receive only cumulative summaries. An advantage for PLVM would therefore provide evidence that the model recovers shared structure from complementary process traces rather than relying only on task-local dynamics or aggregate outcomes.

\section{Results}

\subsection{PowerWash: Predicting Future Navigation Style}
We evaluate PLVM on the main application problem: predicting whether a player will later behave as a Planner or a Hopper in Fire Station from partial Back Garden and Bungalow traces. This is a pre-target-task prediction setting: Fire Station behavior is never observed as input at test time. We focus here on the binary Planner/Hopper task and report the harder three-class version, which includes the intermediate ``Mixed'' group, in the Appendix.

\xhdr{Outcome-Proxy Sanity Check}
Before evaluating process models, we tested whether the Fire Station Planner/Hopper label was simply recoverable from early-task outcomes. At a 20-minute observation cutoff, we computed completion percentage, number of completed subtasks, and subtasks completed per minute for Back Garden and Bungalow. Table~\ref{tab:outcome_proxy_sanity} reports each metric by future Fire Station label, along with the standardized group difference and single-metric AUC. Across these simple outcome measures, future Planners and Hoppers show only weak separation. Back Garden outcomes are nearly indistinguishable across labels, with AUCs of 0.51. Bungalow outcomes show slightly larger differences, but remain weak predictors, with AUCs of 0.55. This suggests that the future navigation-style label is not merely a proxy for early completion or productivity.

\begin{table}[t]
\vspace{-8pt}
\centering
\caption{
Outcome-proxy sanity check after 20 minutes of game play. 
Std. diff. is computed as the Hopper--Planner mean difference divided by the pooled standard deviation.
}
\label{tab:outcome_proxy_sanity}
\begin{tabular}{lcccc}
\toprule
Metric & Planner & Hopper & Std. diff. & AUC \\
\midrule
Back Garden completion (\%)      & $41.5 \pm 19.2$ & $43.3 \pm 23.1$ & $0.08$ & $0.51$ \\
Back Garden subtasks completed   & $47.9 \pm 17.8$ & $49.8 \pm 22.6$ & $0.09$ & $0.51$ \\
Back Garden subtasks/min         & $2.39 \pm 0.89$ & $2.49 \pm 1.13$ & $0.09$ & $0.51$ \\
Bungalow completion (\%)         & $38.6 \pm 20.2$ & $44.4 \pm 24.5$ & $0.26$ & $0.55$ \\
Bungalow subtasks completed      & $22.0 \pm 13.2$ & $24.6 \pm 14.5$ & $0.19$ & $0.55$ \\
Bungalow subtasks/min            & $1.10 \pm 0.66$ & $1.23 \pm 0.73$ & $0.19$ & $0.55$ \\
\bottomrule
\end{tabular}
\end{table}

\begin{figure}[t]
    \centering

    \begin{subfigure}[t]{0.32\linewidth}
        \centering
        \includegraphics[width=\linewidth]{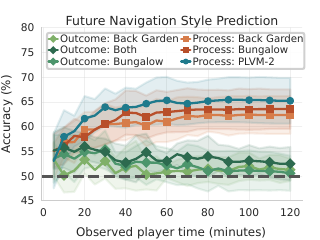}
        \caption{Powerwash.}
        \label{fig:powerwash_plvm_performance}
    \end{subfigure}
    \hfill
    \begin{subfigure}[t]{0.32\linewidth}
        \centering
        \includegraphics[width=\linewidth]{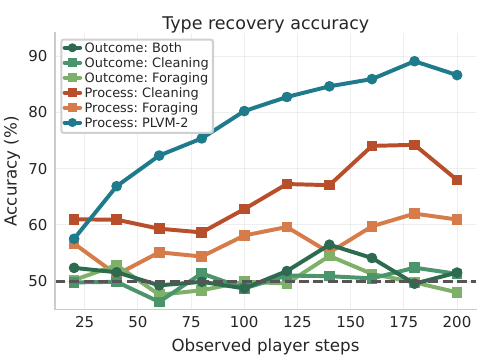}
        \caption{Simulations.}
        \label{fig:plvm_performance_simulation}
    \end{subfigure}
    \hfill
    \begin{subfigure}[t]{0.24\linewidth}
        \centering
        \includegraphics[width=\linewidth]{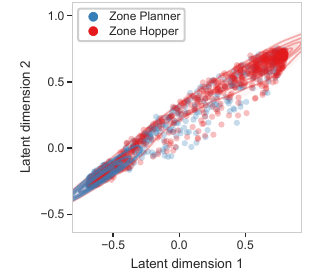}
        \caption{Latent representation.}
        \label{fig:powerwash_latent}
    \end{subfigure}

    \caption{
   Behavioral style prediction.
(a) In PowerWash, models predict a player's Fire Station strategy.
(b) In simulations, models predict ground-truth behavioral type.
(c) A 2-D PLVM latent representation for a representative PowerWash model. Each point is a player. The groups overlap, as expected for a noisy future behavioral strategy, but show a visible shift in latent space.
    }
    \label{fig:powerwash_results_and_latent}
    \vspace{-8pt}
\end{figure}

\xhdr{Cross-Task Prediction Performance}
We next evaluated whether partial Back Garden and Bungalow traces predict a player's later Fire Station navigation style. This is a pre-target-task prediction setting: models do not observe Fire Station behavior at test time, but must predict labels derived from its full trace. Figure~\ref{fig:powerwash_plvm_performance} shows accuracy across observation cutoffs, and Table~\ref{tab:powerwash_summary} summarizes representative time points. Outcome baselines remain close to chance after 60--120 minutes, while single-task process models perform better, indicating that ordered traces contain signal beyond aggregate summaries. PLVM performs best across cutoffs, reaching $65.3\%$ accuracy at 60 minutes and $65.2\%$ at 120, compared with $62.9\%$ and $63.6\%$ for the best single-task process baseline. This pattern supports the main claim: future navigation style is only weakly captured by outcomes, partly visible in individual process traces, and better predicted when process evidence is fused across tasks.

\begin{table}[t]
\vspace{-10pt}
\centering
\caption{
Future navigation-style prediction at representative observation cutoffs; accuracy is mean $\pm$ standard deviation across cross-validation folds.
}
\label{tab:powerwash_summary}
\small
\begin{tabular}{llcccc}
\toprule
Model family  & 20 min Acc. (\%) & 60 min Acc. (\%) & 120 min Acc. (\%) & AUC at 120 min \\
\midrule

Best Outcome  
& $56.0 \pm 2.8$ 
& $53.0 \pm 1.6$ 
& $52.5 \pm 3.4$ 
& $0.54 \pm 0.02$ \\

Best Single-task (Bungalow) 
& $57.9 \pm 3.0$ 
& $62.9 \pm 4.1$ 
& $63.6 \pm 4.0$ 
& $0.62 \pm 0.03$ \\

PLVM 
& $\mathbf{61.6 \pm 4.6}$ 
& $\mathbf{65.3 \pm 4.7}$ 
& $\mathbf{65.2 \pm 4.6}$ 
& $\mathbf{0.64 \pm 0.04}$ \\
\bottomrule
\end{tabular}
\vspace{-8pt}
\end{table}

\xhdr{Latent Representation Analysis}
To examine the learned person-level representation, we visualized the two-dimensional PLVM latent space for a representative PowerWash model in Figure~\ref{fig:powerwash_latent}. Each point represents a player embedding obtained by fusing partial Back Garden and Bungalow traces, colored by the true Fire Station navigation-style label. The labels overlap substantially, consistent with future navigation style being a noisy behavioral phenotype rather than a perfectly separable trait. However, Zone Hoppers are shifted toward the upper-right region of the latent space, while Zone Planners are more concentrated toward the lower-left and middle regions. This suggests that PLVM learns predictive structure relevant to future navigation style.

\vspace{-8pt}
\subsection{Simulation: Mechanism Check with Known Latent Phenotypes}

Because the true latent process is unobserved in PowerWash, we use simulations as a mechanism check: agents have known behavioral types, and paired tasks reveal complementary evidence about those types. Figure~\ref{fig:plvm_performance_simulation} shows validation accuracy as a function of observed agent steps. PLVM receives matched partial windows from both foraging and cleaning, single-task process models receive one task, and outcome baselines receive only cumulative summaries such as completed targets, reward, or cleaned salience. PLVM performs best across cutoffs, while outcome baselines remain close to chance. Single-task process models improve with more observation but remain below PLVM because each task primarily exposes one component of the phenotype: foraging emphasizes opportunistic reward seeking, while cleaning emphasizes patch persistence. We further varied the relationship between simulated tasks; full results are in Appendix Table~\ref{tab:simulation_complementarity}. PLVM helps most when tasks are complementary, improving from 68.0\% for the best single-task model to 86.6\%, and when each task contains weak but shared signal, reaching 74.8\% versus 56.6\%. It provides no benefit when tasks are redundant, where the best single-task model already reaches 89.4\%. These results show that cross-task fusion is useful when tasks provide complementary or noisy views of the same latent process, but not when one task already reveals the relevant structure.

\section{Discussion \& Conclusion}
We introduced PLVM for early cross-task behavioral strategy prediction: using partial traces from earlier tasks to infer how an individual will organize behavior in a future task. Our results support the central claim that future behavioral strategy can be predicted from partial process traces in earlier tasks, but is only weakly captured by aggregate outcomes. In PowerWash, early outcome summaries such as completion, completed subtasks, and productivity provided little separation between players who later behaved as Zone Planners or Zone Hoppers. Single-task process models improved prediction, showing that detailed traces contain signal beyond outcomes, but PLVM performed best by fusing evidence across Back Garden and Bungalow. In simulation, where the latent phenotype is known, the same pattern emerged when different tasks revealed complementary aspects of a shared behavioral process. This pattern clarifies the value of cross-task process modeling. A player's tendency to persist locally or switch broadly may be only partially visible in any single level and may not be reflected in completion or productivity. PLVM is useful in this intermediate setting: each task provides a partial and noisy view of the person, and cross-task fusion combines these views into a representation that better predicts future behavior. The simulation explains why this should happen. Foraging and cleaning each reveal different aspects of the latent phenotype: foraging more strongly exposes opportunistic reward seeking, whereas cleaning more strongly exposes patch persistence. Because the labels depend on the joint configuration of these latent drives, neither task alone fully identifies the phenotype. The PowerWash results show an analogous pattern in real human telemetry, where earlier Back Garden and Bungalow traces contain complementary information about later Fire Station navigation style. For adaptive systems, this matters because useful personalization often has to occur before the target behavior is fully observed.

\xhdr{Broader Impact}
Although the modeling framework proposed in this work has positive potential, we recognize that it could be misused to cause harm, such as unethical behavioral profiling, surveillance, or identification. In the present work, this risk is limited: our simulation uses synthetic agents, and the PowerWash analysis is conducted on an existing research dataset to study aggregate modeling performance rather than to make consequential decisions about individual players. Nevertheless, future applications of PLVM to real-world behavioral data should require clear consent, transparent data collection, privacy protections, and limits on downstream use.

\xhdr{Limitations and Future Work}
PLVM is expected to help when tasks share transferable behavioral structure and no single observed task fully reveals the target behavior. It may provide little benefit when tasks are unrelated, behavior is unstable, or outcomes are already sufficient. Our simulation validates the mechanism under known latent structure, while the PowerWash labels are operational measures of spatial switching rather than externally validated psychological traits; the results should therefore be interpreted as evidence for transferable process structure in these settings, not as a universal guarantee. Future work will test PLVM in other interactive domains, study when fusion helps, fails, or causes negative transfer, and evaluate whether inferred strategies improve downstream interventions and align with external validation measures.

\bibliographystyle{plain}
\bibliography{references.bib}

\newpage
\appendix

\section{Appendix}

This appendix provides additional details needed to reproduce and interpret the experiments. 

\section{PowerWash Dataset and Label Construction}
\label{app:powerwash_labels}

We use three career-mode tasks from the PowerWash Simulator dataset: Back Garden, Bungalow, and Fire Station. 
Back Garden and Bungalow are used as observed source tasks, and Fire Station is used as the held-out future target task. 
All PowerWash models are evaluated on players with usable source-task traces and a valid Fire Station navigation-style label.

\subsection{Cohort Construction}
\label{app:powerwash_cohort}

The PowerWash Simulator dataset contains 11,080 unique players. 
We focus on the career-mode sequence Back Garden $\rightarrow$ Bungalow $\rightarrow$ Fire Station because these tasks form a consistent progression, share the structure of spatial cleaning, and provide substantial within-player overlap. 
Table~\ref{tab:powerwash_funnel} summarizes the progression funnel.

\begin{table}[h]
\centering
\caption{
PowerWash career-mode progression funnel. Percentages are relative to the full dataset of 11,080 unique players.
}
\label{tab:powerwash_funnel}
\small
\begin{tabular}{lcc}
\toprule
Milestone & Players & \% of total \\
\midrule
Total players in dataset & 11,080 & 100.0 \\
Started any career job & 10,347 & 93.4 \\
Completed Back Garden & 8,044 & 72.6 \\
Completed Bungalow & 6,618 & 59.7 \\
Completed Fire Station & 3,846 & 34.7 \\
Completed all three tasks & 3,794 & 34.2 \\
\bottomrule
\end{tabular}
\end{table}

The three-task cohort represents players who progressed substantially through career mode, which is necessary for within-player cross-task prediction. 
Importantly, this cohort is not fragmented: over 98\% of players who completed Fire Station had also completed both Back Garden and Bungalow. 
Only 16 Fire Station completers lacked one of the two earlier completed tasks, likely reflecting edge cases such as FreePlay leakage or logging anomalies.

\begin{table}[h]
\centering
\caption{Cross-task overlap among the PowerWash tasks used in this study.}
\label{tab:powerwash_overlap}
\small
\begin{tabular}{lc}
\toprule
Task pair & Players in common \\
\midrule
Back Garden + Bungalow & 6,589 \\
Back Garden + Fire Station & 3,814 \\
Bungalow + Fire Station & 3,819 \\
\bottomrule
\end{tabular}
\end{table}

For task-specific encoder training, we retained task attempts completed within two hours to remove unusually long or interrupted sessions. 
This yielded 6,012 eligible Back Garden players and 6,061 eligible Bungalow players. 
These are task-specific cohorts and are not required to contain exactly the same players.

For the downstream Fire Station prediction task, we required a scorable and sufficiently complete Fire Station attempt so that the Planner/Hopper label was reliable. 
Of 3,569 players with Fire Station attempt files, 2,541 had at least one scorable attempt with subtask-completion events, and 2,499 met the label criterion of at least 90\% completion, corresponding to at least 165 of 183 unique subtasks completed. 
Requiring these players to also have Back Garden and Bungalow source-task data produced the final PLVM cohort of 2,187 players.

\begin{table}[h]
\centering
\caption{Cohort construction for encoder training and downstream Fire Station prediction.}
\label{tab:powerwash_model_cohorts}
\small
\begin{tabular}{lc}
\toprule
Cohort/filter & Players \\
\midrule
Eligible Back Garden players for encoder training & 6,012 \\
Eligible Bungalow players for encoder training & 6,061 \\
Players with Fire Station attempt files & 3,569 \\
Players with scorable Fire Station subtask-completion events & 2,541 \\
Players meeting Fire Station label criteria & 2,499 \\
Final PLVM cohort with source-task data and Fire Station label & 2,187 \\
\bottomrule
\end{tabular}
\end{table}

These filters ensure that source-task representations are learned from usable traces and that downstream labels are computed from sufficiently complete Fire Station behavior. 
They are intended to support reliable within-player cross-task prediction rather than to select for a desired model outcome.

\subsection{Fire Station Label Construction}
\label{app:firestation_label_construction}

Fire Station navigation-style labels are constructed from the ordered sequence of completed subtasks. 
We cluster Fire Station subtask locations into $K=6$ spatial zones and convert each player's ordered completions into a zone sequence. 
For player $i$, we compute four statistics:
\begin{itemize}
    \item zone switch rate $a_i$, the fraction of consecutive completions that move to a different zone;
    \item return rate $r_i$, the tendency to revisit previously visited zones;
    \item mean zone run length $\ell_i$, the average number of consecutive completions within the same zone;
    \item mean zone jump distance $d_i$, the average spatial distance between successive completed zones.
\end{itemize}

The navigation-style score is
\[
S_i = z(a_i) + z(r_i) - z(\ell_i) + z(d_i),
\]
where $z(\cdot)$ denotes standardization across players. 
Players in the bottom $35\%$ of $S_i$ are labeled \emph{Zone Planners}, players in the top $35\%$ are labeled \emph{Zone Hoppers}, and the middle $30\%$ are treated as \emph{Mixed}/ambiguous and excluded from the primary binary analysis.

\section{Model Inputs, Hyperparameters, and Encoder Performance}
\label{app:model_details}

All models are evaluated under matched observation cutoffs. 
At each cutoff, outcome baselines compute summaries only from the observed prefix, single-task process models encode one observed-task prefix, and PLVM fuses prefixes from both observed tasks. 
The held-out Fire Station trace is never provided as input at test time.

\begin{table}[h]
\centering
\caption{
Inputs available to each model family.
Outcome baselines use aggregate summaries, whereas process models use ordered behavioral traces.
}
\label{tab:model_inputs}
\small
\begin{tabular}{llp{7.2cm}}
\toprule
Experiment & Model family & Inputs \\
\midrule
Simulation 
& Outcome baseline 
& Cumulative summaries up to cutoff, including completion count/percentage, total reward, high-value target counts/reward, and salience totals. \\

Simulation 
& Single-task process 
& Ordered behavioral trace from one simulated task only, encoded by the corresponding task-specific transformer. \\

Simulation 
& PLVM 
& Ordered behavioral traces from both simulated tasks at matched cutoffs, encoded by frozen task-specific transformers and fused downstream. \\

PowerWash 
& Outcome baseline 
& Aggregate summaries from the observed prefix only, including completion percentage, completed subtasks, subtasks per minute, recent-window counts, trend/slope, zero-rate, and participant running mean. \\

PowerWash 
& Single-task process 
& Ordered subtask-completion trace from one observed PowerWash task only, encoded by the corresponding task-specific transformer. \\

PowerWash 
& PLVM 
& Ordered subtask-completion traces from Back Garden and Bungalow at matched cutoffs, encoded by frozen task-specific transformers and fused downstream. \\
\bottomrule
\end{tabular}
\end{table}

For the main binary PowerWash task, we report validation accuracy averaged across 10 cross-validation folds. 
For sanity checks, we also report single-metric AUCs to quantify whether simple outcome summaries separate future Planners and Hoppers.

\subsection{Model Hyperparameters}
\label{app:hyperparameters}

\begin{table}[h]
\centering
\caption{Model hyperparameters used in the reported experiments.}
\label{tab:hyperparameters}
\small
\begin{tabular}{lccccc}
\toprule
Model & Hidden size & Layers & Heads & Context length & Learning rate \\
\midrule
Simulation transformer & 256 & 2 & 4 & 200 steps & 0.001 \\
PowerWash transformer & 32 & 2 & 4 & 120 events & 0.002, 0.01 \\
Simulation PLVM & 256 encoder / 32 decoder & 2 decoder & -- & -- & 0.0001 \\
PowerWash PLVM & 32 encoder / 32 decoder & 3 decoder & -- & -- & 0.0001 \\
Outcome baselines & -- & -- & -- & -- & tuned by validation \\
\bottomrule
\end{tabular}
\end{table}

Task-specific transformers are trained independently using within-task next-event prediction objectives and are frozen before training the PLVM fusion and decoder modules. 
For simulation, the next-event target is the next action or target choice. 
For PowerWash, the next-event target is the player's position change by the next subtask-completion event. 
All downstream process models use the same frozen task-specific encoders, so comparisons between single-task process models and PLVM differ only in whether one or multiple task embeddings are used.

\subsection{Task-Specific Encoder Performance}
\label{app:encoder_performance}

Task-specific encoders are not the final prediction models; they are used to learn within-task process representations before downstream fusion. 
In simulation, the task-specific transformers achieved held-out next-action prediction accuracy of approximately $85$--$88\%$, indicating that they learned the within-task target-selection dynamics. 
In PowerWash, the encoders predicted the player's position change by the next subtask-completion event. 
Table~\ref{tab:encoder_performance} reports held-out performance for both domains.

\begin{table}[h]
\centering
\caption{
Held-out performance of task-specific process encoders.
Simulation encoders are evaluated by next-action prediction accuracy; PowerWash encoders are evaluated by RMSE for position change to the next subtask-completion event.
}
\label{tab:encoder_performance}
\small
\begin{tabular}{llccc}
\toprule
Domain/task & Prediction target & Metric 1 & Metric 2 & Metric 3 \\
\midrule
Simulation foraging/cleaning 
& Next action or target choice 
& \multicolumn{3}{c}{$85$--$88\%$ accuracy} \\

PowerWash Back Garden 
& $\Delta x,\Delta y,\Delta z$ to next subtask 
& $2.70$ 
& $0.39$ 
& $2.26$ \\

PowerWash Bungalow 
& $\Delta x,\Delta y,\Delta z$ to next subtask 
& $3.46$ 
& $0.72$ 
& $2.78$ \\
\bottomrule
\end{tabular}
\end{table}

\section{Additional Results}
\label{app:additional_results}

\subsection{Simulation Complementarity Analysis}
\label{app:complementarity_analysis}

We ran an additional simulation analysis to test when cross-task fusion should help. 
The main simulation uses complementary tasks: foraging more strongly expresses opportunistic reward seeking, while cleaning more strongly expresses patch persistence. 
We compare this setting with two alternatives. 
In the redundant condition, both tasks reveal similar information, so a strong single-task model should already recover the label. 
In the weak-but-shared condition, each task contains noisy evidence about the same latent process, so fusion can help by aggregating weak signals.

\begin{table}[h]
\centering
\caption{
Simulation complementarity analysis.
PLVM helps most when tasks provide complementary or individually weak evidence about a shared latent process, but provides little benefit when one task is already sufficient.
}
\label{tab:simulation_complementarity}
\small
\begin{tabular}{lccc}
\toprule
Task relationship & Best single-task & PLVM & PLVM gain \\
\midrule
Complementary tasks & 68.0 & 86.6 & +18.6 \\
Redundant tasks & 89.4 & 88.4 & -1.0 \\
Weak but shared signal & 56.6 & 74.8 & +18.2 \\
\bottomrule
\end{tabular}
\end{table}

The results support the intended mechanism. 
PLVM gives a large gain when tasks reveal complementary aspects of the phenotype and when each task contains weak but shared signal. 
When tasks are redundant, the best single-task model already performs well and fusion provides no additional benefit. 
This clarifies the scope of PLVM: cross-task fusion is useful when tasks provide different or noisy views of the same latent process, but it is not expected to help when one task already reveals the relevant structure.

\subsection{Three-Class PowerWash Prediction}
\label{app:three_class_powerwash}

\begin{figure}
    \centering
    \includegraphics[width=0.8\linewidth]{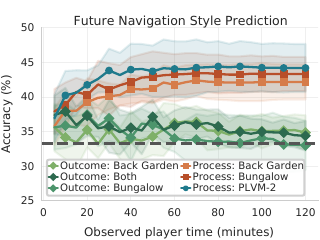}
    \caption{
    Three-class Fire Station navigation-style prediction, including Zone Planner, Mixed, and Zone Hopper labels.
    Performance decreases relative to the binary task because the Mixed group is less behaviorally separable.
    }
    \label{fig:three_class_powerwash}
\end{figure}

The main PowerWash analysis uses the binary Planner/Hopper task, excluding the ambiguous middle $30\%$. 
We also evaluate a three-class setting that includes the Mixed group. 
As expected, this task is harder because the Mixed group is behaviorally intermediate rather than clearly separated. 
Figure~\ref{fig:three_class_powerwash} shows that performance decreases for all models, but PLVM retains an advantage over single-task process models, while outcome models remain close to chance.

\subsection{PLVM Latent Dimension Sensitivity}
\label{app:latent_dim_sensitivity}

\begin{figure}[h]
    \centering
    \includegraphics[width=0.8\linewidth]{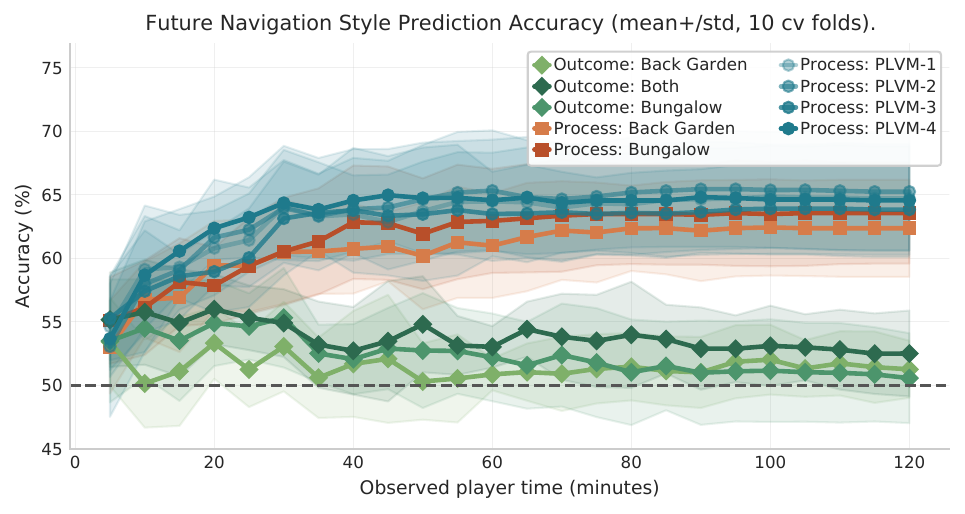}
    \caption{
    Latent-dimension sensitivity for the binary PowerWash Planner/Hopper prediction task.
    PLVM variants with low-dimensional latent states perform similarly, with PLVM-2 providing the strongest and most interpretable overall model.
    }
    \label{fig:plvm_dim_sensitivity_binary}
\end{figure}

\begin{figure}[h]
    \centering
    \includegraphics[width=0.8\linewidth]{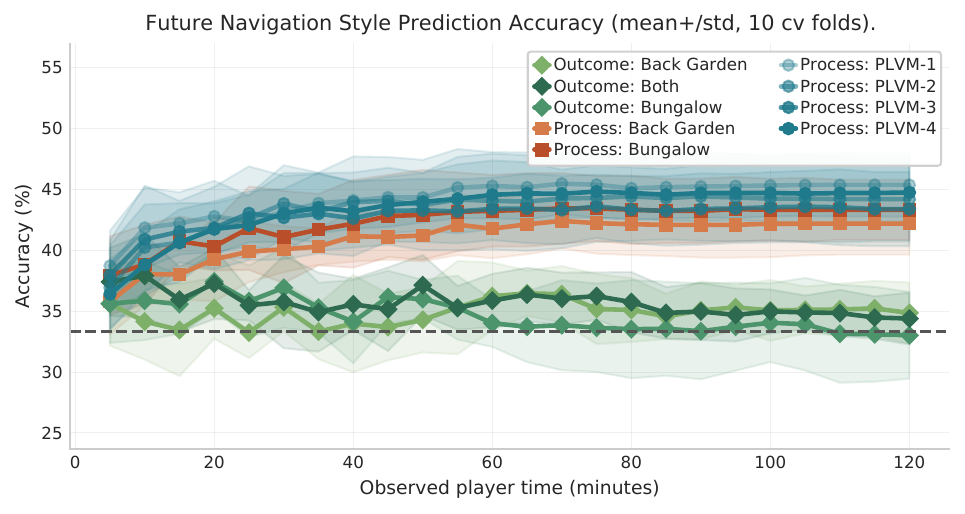}
    \caption{
    Latent-dimension sensitivity for the three-class PowerWash Planner/Mixed/Hopper prediction task.
    The three-class task is harder, but PLVM variants remain competitive with each other and outperform outcome-only baselines.
    }
    \label{fig:plvm_dim_sensitivity_three_class}
\end{figure}

We evaluated PLVM variants with different latent dimensionalities to test whether performance depended on a specific choice of latent size. 
Figures~\ref{fig:plvm_dim_sensitivity_binary} and~\ref{fig:plvm_dim_sensitivity_three_class} show results for the binary Planner/Hopper task and the harder three-class Planner/Mixed/Hopper task. 
Across both settings, low-dimensional PLVM variants perform competitively, and the two-dimensional model provides the best overall tradeoff between accuracy and interpretability. 
We therefore use PLVM-2 in the main analyses: it is simple enough to visualize directly, while still matching or exceeding the performance of higher-dimensional variants across most observation cutoffs.

\section{Simulation Details}
\label{app:simulation_details}

The simulation uses paired grid-world tasks to test whether cross-task fusion can recover known latent behavioral phenotypes. 
Each agent completes both a foraging task and a cleaning task. 
Agents differ only in target-selection behavior; movement to the selected target follows shortest-path navigation.

\begin{table}[h]
\centering
\caption{Simulation parameters used in the paired-task grid-world environment.}
\label{tab:simulation_params}
\small
\begin{tabular}{ll}
\toprule
Parameter & Value \\
\midrule
Number of agents & 8,000 \\
Grid size & $20 \times 20$ \\
Foraging targets & 32 \\
Cleaning targets & 48 \\
High-value food items & 2 \\
Low-value food reward & 1.0 \\
High-value food reward & 3.0 \\
Cleaning target reward & 1.0 \\
Movement cost & 0.01 \\
$\beta_P^{\mathrm{forage}}$ & 0.20 \\
$\beta_O^{\mathrm{forage}}$ & 2.0 \\
$\beta_P^{\mathrm{clean}}$ & 1.8 \\
$\beta_O^{\mathrm{clean}}$ & 0.6 \\
$\beta_D$ & 0.10 \\
Noise level & 0.5 \\
Softmax temperature & 1.0 \\
Observation cutoffs & 20, 40, 60, 80, 100, 120, 140, 160, 180, 200 steps \\
\bottomrule
\end{tabular}
\end{table}

Latent variables are sampled from class-conditional mixtures:
\[
\mathrm{Type\ A}: 
\tfrac{1}{2}\mathcal{N}((0.8,0.8), 0.35^2 I) +
\tfrac{1}{2}\mathcal{N}((-0.8,-0.8), 0.35^2 I),
\]
\[
\mathrm{Type\ B}: 
\tfrac{1}{2}\mathcal{N}((0.8,-0.8), 0.35^2 I) +
\tfrac{1}{2}\mathcal{N}((-0.8,0.8), 0.35^2 I).
\]
Type A agents have same-sign persistence and opportunism drives, whereas Type B agents have opposite-sign drives.



\section{Computation Details}
\label{app:compute}

All experiments were run on a computing cluster with 40 CPU cores, an Intel Xeon Gold 6148 CPU at 2.40GHz, one NVIDIA Tesla V100 SXM2 GPU with 32GB memory, and up to 80GB system memory. 
Experiments used Python 3.10, PyTorch for model training, and NumPy/pandas for data processing.

\clearpage

\newpage

\end{document}